\documentclass[10pt,twocolumn,letterpaper]{article}
\usepackage{arxiv}
\usepackage{microtype}
\usepackage{tabularx}
\usepackage{multirow}
\usepackage{enumitem}
\usepackage{colortbl}
\usepackage{placeins}
\newcommand{\ind}{\mathbf{1}}
\newcommand{\pairacc}{\operatorname{PairAcc}}
\newcommand{\acc}{\operatorname{Acc}}
\newcommand{\blindicon}{\raisebox{-0.10em}{\includegraphics[height=0.85em]{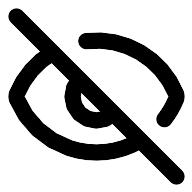}}}

\definecolor{linkblue}{rgb}{0.21,0.49,0.74}
\usepackage[breaklinks,colorlinks,allcolors=linkblue]{hyperref}
\hypersetup{
  pdftitle={TwinPast: Benchmarking Historical Evidence Use in Streaming Video Understanding with Evidence Contrastive Video Pairs},
  pdfauthor={Haotian Yang, Yong Chen, Cheng Peng, Haobo Wang, Zhongle Xie, Xuanru Li, Gang Chen},
  pdfsubject={Historical evidence use in streaming video understanding},
  pdfkeywords={TwinPast, streaming video understanding, historical evidence, video benchmark}
}
\newcommand{\projecturl}{https://yh1zzzzzzz.github.io/twinpast/}
\newcommand{\dataseturl}{https://huggingface.co/datasets/younght/TwinPast}
\newcommand{\codeurl}{https://github.com/Yh1zzzzzzz/TwinPast_code}
\newcommand{\resourcelinks}{%
  \href{\dataseturl}{%
    \raisebox{-0.22em}{\includegraphics[height=1.3em]{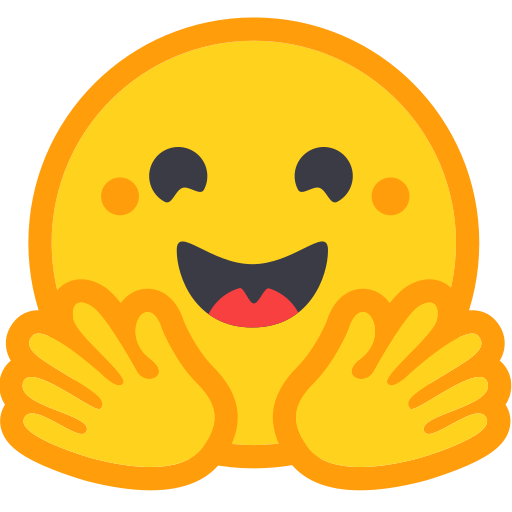}}\hspace{0.45em}%
    \expandafter\nolinkurl\expandafter{\dataseturl}}\\[1pt]
  \href{\codeurl}{%
    \raisebox{-0.22em}{\includegraphics[height=1.3em]{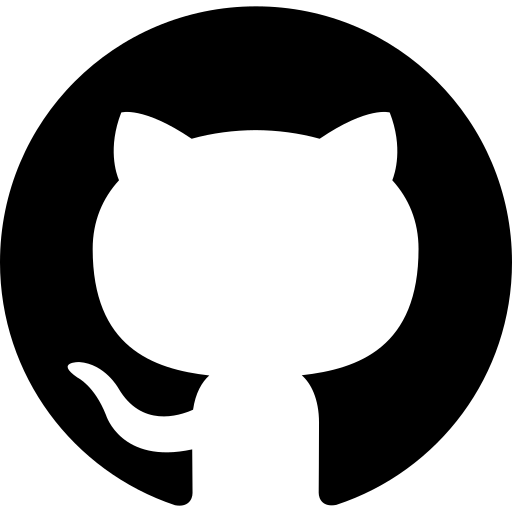}}\hspace{0.45em}%
    \expandafter\nolinkurl\expandafter{\codeurl}}%
}
\title{TwinPast: Benchmarking Historical Evidence Use in Streaming Video Understanding with Evidence Contrastive Video Pairs}
\author{%
  Haotian Yang\textsuperscript{1}\quad Yong Chen\textsuperscript{1}\quad
  Cheng Peng\textsuperscript{1,*}\quad Haobo Wang\textsuperscript{1}\\[2pt]
  Zhongle Xie\textsuperscript{1}\quad Xuanru Li\textsuperscript{2}\quad
  Gang Chen\textsuperscript{3}\\[4pt]
  {\small\textsuperscript{1}School of Software Technology, Zhejiang University\quad
  \textsuperscript{2}HeyGen\quad\textsuperscript{3}Zhejiang University}\\[-1pt]
  {\small\hypersetup{urlcolor=black}%
    \{\href{mailto:yanghaotian45@zju.edu.cn}{yanghaotian45},%
    \href{mailto:chengchng@zju.edu.cn}{chengchng},%
    \href{mailto:wanghaobo@zju.edu.cn}{wanghaobo},%
    \href{mailto:xiezl@zju.edu.cn}{xiezl},%
    \href{mailto:cg@zju.edu.cn}{cg}\}@zju.edu.cn}\\[-2pt]
  {\small\hypersetup{urlcolor=black}%
    \href{mailto:xuanru.li@heygen.com}{xuanru.li@heygen.com}}
}
\date{}
\makeatletter
\g@addto@macro\@maketitle{\begingroup
\captionsetup{hypcap=false}
\centering
\includegraphics[width=0.87\textwidth,trim=0bp 13bp 0bp 7bp,clip]{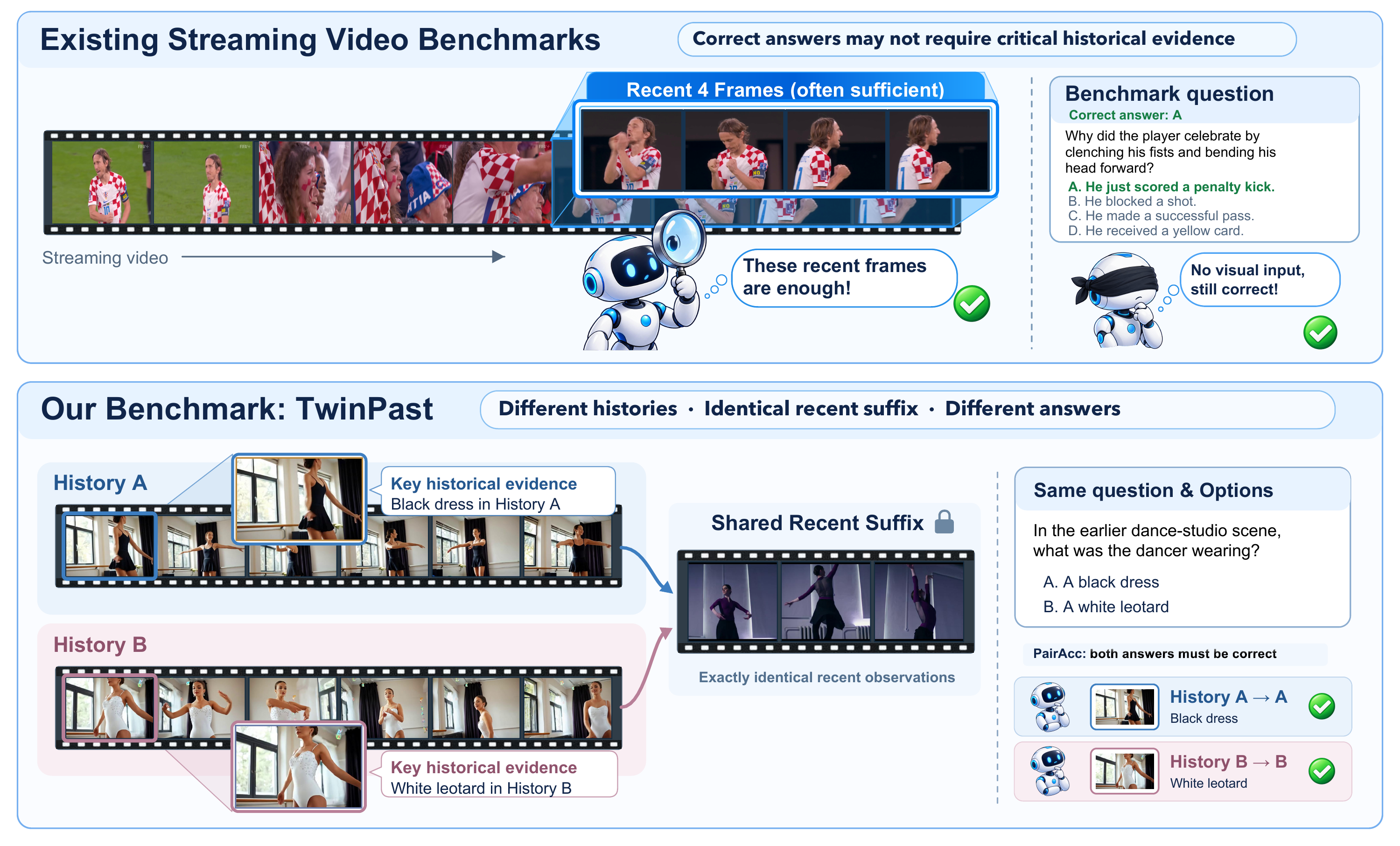}
\captionof{figure}{\textbf{TwinPast overview.} Existing benchmarks can reward answers without historical evidence. TwinPast fixes recent context, questions, and options while changing history and correct answers; PairAcc requires both branches correct.}
\label{fig:protocol}
\label{fig:teaser}
\par
\endgroup
}
\makeatother
\begin{document}
\maketitle
\begingroup
\renewcommand{\thefootnote}{}
\makeatletter
\long\def\@makefntext#1{\noindent#1}
\def\Hy@footnote@currentHref{corresponding-author}
\makeatother
\footnotetext{\textsuperscript{*}Corresponding author.}
\endgroup
\begin{abstract}
High accuracy on streaming video benchmarks does not establish whether a model uses visual history. Our review of OVO-Bench and StreamingBench finds that recent frames suffice for 32\% of sampled historical questions, while models also answer many questions correctly from the question and options alone. We introduce \textbf{TwinPast} to test whether answers correctly follow changes in history, using 350 human-reviewed video pairs spanning five historical tasks. Each pair holds the recent video context, question, and ordered options fixed while changing critical historical evidence and the correct answer. Each video is evaluated independently, and PairAcc requires correct answers for both histories. We evaluate 34 model and input configurations. For Qwen3-VL-8B and DeepSeek-Flash, historical inputs improve performance on TwinPast, reversing the recent-frame advantage on existing benchmarks. The strongest evaluated streaming system reaches 49.14\% PairAcc, compared with 66.57--91.43\% for the evaluated general-purpose models. This gap persists across all five tasks. Case studies further show that models select states from the wrong time or contradict recorded event order despite retaining the relevant evidence. These findings highlight the need to connect historical memory and retrieval with reasoning that uses evidence according to the question's temporal references and event relations.
\end{abstract}

\section{Introduction}
Streaming video systems are expected to answer questions about events that are no longer visible: which object appeared first, how an earlier action ended, or how many times it occurred~\cite{sun2025video,fu2026streamflow}. Yet a question about the past can sometimes be answered without observing that past~\cite{cores2025losttimenewtemporal,liu-etal-2024-tempcompass}. SimpleStream's strong recent context results~\cite{shen2026simplebaselinestreamingvideo} raise a question about what these benchmarks measure: how much does a high benchmark score reveal about a system's use of visual history?

We investigate this question on OVO-Bench~\cite{11094230} and StreamingBench~\cite{11463959} by comparing model accuracy under four input settings. Models receive the question and options alone, or with the latest frame, up to four recent frames, or up to 64 uniformly sampled frames. On the same 4,276 questions, Qwen3-VL-8B~\cite{bai2025qwen3vltechnicalreport} and DeepSeek-Flash achieve 40.83\% and 49.60\% accuracy using only the question and options, without any visual input. Manual review of 1,200 questions also finds that recent frames alone suffice for 128 of 400 questions sampled specifically from historical tasks. These results show two ways models can answer correctly without earlier visual input: using information already visible in recent frames or relying on the question and options alone. Ordinary accuracy credits both, leaving historical evidence use unresolved~\cite{xiao2024can}.

We introduce \textbf{TwinPast} to test whether answers correctly follow changes in history. Each video pair shares the recent video context, question, and ordered options, while the earlier events and correct answers differ (Fig.~\ref{fig:protocol}). The shared inputs therefore cannot distinguish the two histories. We evaluate each video independently and use \emph{PairAcc}, which \textbf{requires both answers to be correct.} Repeating an answer favored by the shared text or recent scene can succeed on one video but cannot solve the pair. TwinPast contains 350 human-reviewed pairs covering attribute and identity recall, historical states, event order, action counting, and multi-event integration.

Our evaluation of 34 model and input configurations reveals a gap between individual video accuracy and paired performance. Qwen3-VL-8B answers 42\% of individual questions correctly with four recent frames but solves no pairs; uniformly sampling up to 64 frames from the available video raises PairAcc to 34\%. DeepSeek-Flash likewise improves from 3.14\% to 42.57\% PairAcc. Historical input also improves individual accuracy for both models on TwinPast, \textbf{whereas the reported comparisons on OVO-Bench and StreamingBench favor recent frames.} This reversal highlights the value of historical input when evaluation requires distinguishing different pasts under the same recent context.

All evaluated streaming systems fall below general purpose models in paired accuracy. The strongest streaming system reaches 49.14\% PairAcc, compared with 66.57--91.43\% for the general purpose models, and this gap persists across all five historical tasks. Among models evaluated both online and offline, most achieve lower paired accuracy online. The Qwen3.8-Omni family provides a clear example: its realtime variant achieves lower paired accuracy and produces different answers on far fewer pairs than its non-realtime counterpart. These comparisons show that offline historical question answering performance is not consistently matched in realtime settings.

These gaps can arise even when relevant historical evidence remains available. Case analyses show ReKV selecting an initial rather than queried state despite retaining both, and OneStreamer contradicting the event order recorded in its memory. Historical QA therefore requires both retaining evidence and using its temporal references correctly.

To summarize, we make the following contributions: (1) We audit OVO-Bench and StreamingBench and identify correct answers supported by recent views or by questions and options alone. (2) We introduce TwinPast, a benchmark of 350 video pairs that tests responses to different histories under identical recent context and text. (3) We evaluate 34 configurations, quantifying gaps in historical discrimination and identifying repeated answers, incomplete responses to multiple facts, and errors in using retained temporal evidence.

\section{Related Work}
\paragraph{Streaming video understanding benchmarks.}
OVO-Bench~\cite{11094230} covers backward tracing, real-time perception, and forward active responding. StreamingBench~\cite{11463959} evaluates real-time visual, omni-source, and contextual understanding, including proactive output. SPOT-Bench~\cite{chatterjee2026dontpausepredictionmatters} evaluates response timing and coverage throughout a stream; StreamArena~\cite{zhang2026streamarenacontinuousinteractivelonghorizon} studies hour scale perception, historical recall, proactive interaction, and tool use. S-EMBER~\cite{wang2026semberlargescalebenchmarkstreaming} evaluates egocentric memory retrieval with temporal evidence localization.

\paragraph{Streaming video understanding models.}
VideoLLM-online~\cite{chen2024videollm} learns response timing through streaming EOS prediction, while Dispider~\cite{qian2025dispider} separates perception, response decisions, and asynchronous generation. To manage historical context~\cite{song2024moviechat,he2024ma,yao2025timechat}, ReKV~\cite{di2025streaming} retrieves query relevant video KV caches, and StreamForest~\cite{zeng2025streamforest} combines a recent visual window with hierarchical event memory. AURA~\cite{lu2026aura} maintains separate windows for recent video and QA history. OneStreamer~\cite{zeng2026onestreamer} combines recent visual context with timestamped caption memory and learns when to record evidence or respond.

\paragraph{Paired and counterfactual visual evaluation.}
VQA v2~\cite{Goyal_2017_CVPR} pairs complementary images with a shared question to reduce language bias. DyBench~\cite{du2026learningspatiotemporalsensitivityvideo} measures joint correctness on video pairs with altered spatiotemporal dynamics. DualityVidQA~\cite{11698217} pairs original and counterfactually edited videos, while VidPair-Halluc~\cite{huang2026no} controls backgrounds while varying foreground semantics. TwinPast extends paired evaluation~\cite{Thrush_2022_CVPR,zhang2024vinogroundscrutinizinglmmsdense,krojer2025shortcutawarevideoqabenchmarkphysical} to historical evidence use: recent observations, questions, and ordered options remain identical while histories and correct answers differ.

\section{Correct Answers without History}
\label{sec:audit}
Evaluating historical understanding requires questions whose answers depend on earlier events. Our review of OVO-Bench~\cite{11094230} and StreamingBench~\cite{11463959} identifies two ways models can answer correctly without observing those events: recent frames may already contain sufficient visual evidence, while the question and options alone may also lead to correct answers. We examine both through input comparisons and manual evidence review, then use these findings to motivate TwinPast's paired design.

\subsection{High Scores with Little or No Video}
\label{sec:audit_inputs}
Following SimpleStream's recent context baseline~\cite{shen2026simplebaselinestreamingvideo}, we compare DeepSeek-Flash and Qwen3-VL-8B~\cite{bai2025qwen3vltechnicalreport} under four input conditions on the same questions within each benchmark (Table~\ref{tab:input_comparison}). \emph{Query only} uses only the question and options; \emph{1 Frame} and \emph{4 Frames} use the latest frame and up to four recent frames, respectively, while \emph{64 Frames} uniformly samples up to 64 frames available at query time.

\begin{table*}[t]
\centering\small
\caption{Input comparison (\%). OVO Avg equally weights Real-Time/Backward task macros; StreamingBench groups are question weighted.}
\label{tab:input_comparison}
\setlength{\tabcolsep}{3pt}
\begin{tabular}{@{}l*{6}{c}@{\hspace{11pt}}*{4}{c}@{}}
\toprule
 & \multicolumn{6}{c}{OVO-Bench} & \multicolumn{4}{c}{StreamingBench} \\
\cmidrule(lr){2-7}\cmidrule(lr){8-11}
\rule{0pt}{4.5ex}Model / input & EPM & ASI & HLD & \shortstack{Real-Time\\Avg} & \shortstack{Backward\\Avg} & OVO Avg & ACU & MCU & \shortstack{Real-Time\\Avg} & \shortstack{Contextual\\Avg} \\
\midrule
\multicolumn{11}{l}{\textbf{(A) Streaming Systems}} \\
VideoLLM-online-8B~\cite{chen2024videollm} & 22.22 & 18.80 & 12.18 & 20.79 & 17.73 & 19.26 & 27.82 & 30.80 & 35.99 & 29.32 \\
Dispider-7B~\cite{qian2025dispider} & 48.48 & 55.41 & 4.30 & 54.55 & 36.06 & 45.31 & 42.34 & 35.20 & 67.63 & 38.76 \\
MiniCPM-o 4.5~\cite{cui2026minicpm} & 55.20 & 66.90 & 45.70 & 67.30 & 55.90 & 61.60 & 41.94 & 29.20 & 78.20 & 35.54 \\
OneStreamer-4B~\cite{zeng2026onestreamer} & 62.60 & 71.60 & 99.50 & 81.40 & 77.90 & 79.65 & 68.55 & 60.40 & 86.90 & 64.46 \\
StreamForest-7B~\cite{zeng2025streamforest} & 58.92 & 64.86 & 32.26 & 61.20 & 52.02 & 56.61 & 37.90 & 31.20 & 77.26 & 34.54 \\
ReKV-7B~\cite{di2025streaming} & 54.20 & 57.40 & 28.50 & 64.40 & 46.70 & 55.55 & 29.44 & 36.40 & 69.10 & 32.93 \\
MOSS-VL Realtime~\cite{wang2026moss} & 60.60 & 64.50 & 92.40 & 75.90 & 72.60 & 74.25 & 58.82 & 51.25 & 82.90 & 55.02 \\
\midrule
\multicolumn{11}{l}{\textbf{(B) Frame based VLMs}} \\
DeepSeek-Flash 1 Frame & 47.14 & 68.49 & 72.58 & 73.36 & 62.74 & 68.05 & 53.23 & 64.80 & 77.79 & 59.04 \\
DeepSeek-Flash 4 Frames & 54.88 & 69.18 & 68.28 & 76.82 & 64.11 & 70.47 & 63.71 & 57.60 & 80.67 & 60.64 \\
DeepSeek-Flash 64 Frames & 53.54 & 67.81 & 52.69 & 64.36 & 58.01 & 61.18 & 38.31 & 37.60 & 74.27 & 37.95 \\
\addlinespace[3pt]
Qwen3-VL-8B 1 Frame & 49.16 & 51.37 & 46.77 & 67.83 & 49.10 & 58.47 & 43.15 & 58.40 & 73.03 & 50.80 \\
Qwen3-VL-8B 4 Frames & 53.20 & 60.96 & 39.78 & 82.55 & 51.31 & 66.93 & 67.74 & 64.80 & 81.79 & 66.27 \\
Qwen3-VL-8B 64 Frames & 51.18 & 58.22 & 37.63 & 52.91 & 49.01 & 50.96 & 24.19 & 32.40 & 67.71 & 28.31 \\
\midrule
\multicolumn{11}{l}{\textbf{(C) Blind LLMs (no visual input)}} \\
\blindicon\hspace{0.35em}DeepSeek-Flash Query only & 37.04 & 47.95 & 88.71 & 40.14 & 57.90 & 49.02 & 43.15 & 28.80 & 57.18 & 35.94 \\
\blindicon\hspace{0.35em}Qwen3-VL-8B Query only & 35.69 & 50.68 & 69.89 & 30.74 & 52.09 & 41.41 & 27.82 & 28.40 & 46.70 & 28.11 \\
\bottomrule
\end{tabular}
\end{table*}

With only one frame, DeepSeek-Flash reaches 68.05\% OVO Avg and 77.79\% StreamingBench Real-Time Avg; Qwen3-VL-8B reaches 58.47\% and 73.03\%, respectively. For both models, 1 Frame and 4 Frames exceed 64 Frames on every aggregate score across both benchmarks in Table~\ref{tab:input_comparison}.

Query only also achieves substantial accuracy: Qwen3-VL-8B and DeepSeek-Flash answer 40.83\% and 49.60\% correctly on the same 4,276 questions. We next examine the evidence in recent frames, followed by the answers obtainable from text alone.

\subsection{What the Recent Frames Already Reveal}
\label{sec:evidence_audit}
We manually review 1,200 questions with their options and three or four recent frames: 400 per benchmark sampled across task categories, followed by 200 additional historical task questions per benchmark. Sampling is random within categories with proportional quotas; the two samples are disjoint. Historical tasks comprise OVO-Bench's Episodic Memory (\textbf{EPM}) and Action Sequence Identification (\textbf{ASI}) tasks, and StreamingBench's Causal Reasoning (\textbf{CR}), Clips Summarize (\textbf{CS}), Counting (\textbf{CT}), and Event Understanding (\textbf{EU}) tasks. Evidence is sufficient when it uniquely supports the reference answer.

Recent frames suffice for 219/400 OVO questions (54.75\%) and 244/400 StreamingBench questions (61.00\%) in the sample across task categories. The reviewed questions span perception, historical, and contextual tasks. In the additional historical task sample, \textbf{128/400 questions (32.00\%)} are answerable from recent views. Thus, a historical task label does not by itself establish that earlier visual input is necessary. Combining perception and historical tasks into a single score makes it harder to assess historical understanding, since recent frames are often sufficient for perception questions.

\subsection{Answers from Text Alone}
In the sample across task categories, we identify 175 questions whose recent frames lack sufficient answer evidence. Using only the question and options, DeepSeek-Flash correctly answers 83 (47.43\%), and Qwen3-VL-8B answers 64 (36.57\%). In the additional historical task sample, the two models correctly answer 67 (45.58\%) and 65 (44.22\%) of 147 such questions, respectively. Substantial correct answer rates thus persist without visual input. Task level results, examples, and option order controls are provided in the supplement.

All 186 evaluated OVO Hallucination Detection questions have ``Unable to answer'' as the correct response. Always choosing this option yields 100\% accuracy without video, so this subset alone does not test discrimination between answerable and unanswerable questions.

\paragraph{Making historical evidence necessary.}
\label{sec:requirements}
TwinPast addresses these ambiguities by changing the correct answer across histories while fixing the recent frames, question, and options. Repeating an answer supported only by the shared inputs cannot solve both branches.

\section{TwinPast: Design and Construction}
\label{sec:benchmark}
\subsection{Paired Histories with Shared Recent Context}
For video pair $i$, let
\begin{equation}
V_i^A=P_i\oplus H_i^A\oplus R_i,\qquad
V_i^B=P_i\oplus H_i^B\oplus R_i,
\label{eq:pair}
\end{equation}
where $\oplus$ denotes temporal concatenation, $P_i$ is an optional common prefix, $H_i^A,H_i^B$ are different histories, and $R_i$ is the shared recent suffix. Both branches use the same question $q_i$, ordered options $o_i$, and query time $T_i$, with different answers $y_i^A\ne y_i^B$. Branch identifiers and reference answers are excluded from prompts. Evaluation uses separate conversations and memory states for the two branches.

\begin{figure*}[t]
\centering
\includegraphics[width=\textwidth,trim=0bp 34bp 0bp 4bp,clip]{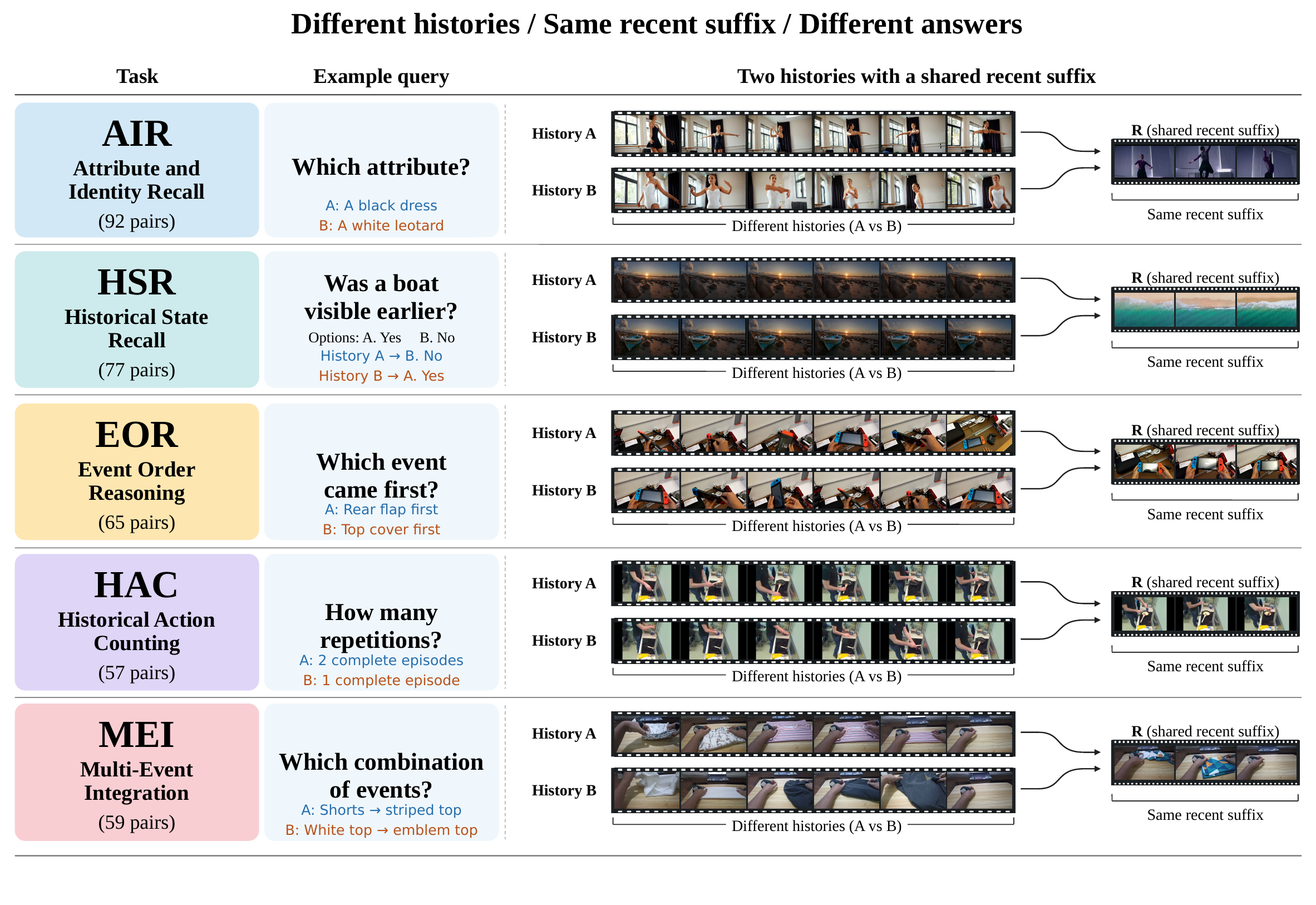}
\caption{TwinPast task examples: contrasting histories yield different answers to the same question with identical recent frames.}
\label{fig:task_overview}
\end{figure*}

\subsection{Construction and Quality Checks}
We construct TwinPast from MVP~\cite{krojer2025shortcutawarevideoqabenchmarkphysical}, DyBench~\cite{du2026learningspatiotemporalsensitivityvideo}, ReCo-Data~\cite{zhang2025regionconstraintincontextgenerationinstructional}, DualityVidQA~\cite{11698217}, VidPair-Halluc~\cite{huang2026no}, and Perception Test~\cite{patraucean2023perceptiontestdiagnosticbenchmark}. We select or compose histories to contrast the queried attributes, states, events, or temporal arrangements while matching backgrounds, viewpoints, and unrelated objects as closely as the source material permits~\cite{Kuehne_2014_CVPR,eidon2026trackerpov,benshabat2023ikeaasmdatasetunderstanding,Wang_2023_ICCV}. We choose a shared suffix visually consistent with both preceding histories in scene appearance, camera viewpoint, object arrangement, and ongoing activities; its frames are identical across branches.

Automatic checks and AI assisted review are followed by human inspection of every pair, English question, and answer key. Review checks whether the historical differences support the intended answers, examines unrelated visual differences, and assesses continuity at the shared suffix transitions. All 350 pairs passed human review. We retain source information and annotate relevant events and time intervals where available.

\subsection{Task Coverage}
TwinPast contains five historical task groups, each testing a different aspect of historical video understanding (Fig.~\ref{fig:task_overview}). These include temporal reasoning~\cite{xiao2021next,li2024vitatecs,cai2024temporalbench} and action repetition counting~\cite{Dwibedi_2020_CVPR}, evaluated here through paired historical contrasts.
\textbf{AIR: Attribute and Identity Recall.} Retrieve an entity's identity or earlier visual, spatial, or motion attributes. Some questions identify the entity through its appearance order or actions, requiring the correct entity to be located first.
\textbf{HSR: Historical State Recall.} Recover an object's condition or configuration at a specified past moment. Across multiple state changes, the model must select the requested state rather than merely recognize states that appeared.
\textbf{EOR: Event Order Reasoning.} Recover the temporal order of past events. Histories may contain the same events in different orders, requiring their sequence rather than merely their occurrence to be recognized.
\textbf{HAC: Historical Action Counting.} Count past action occurrences or affected objects. These questions require tracking individual occurrences across history rather than simply recognizing an activity.
\textbf{MEI: Multi-Event Integration.} Combine or compare multiple historical facts, such as outcomes of two demonstrations or successive attribute changes. Correct answers require integrating evidence across events rather than recalling a single event.
\subsection{Evaluation Metrics}
For $N$ video pairs and predictions $\hat y_i^b$, define
\begin{align}
\acc&=\frac{1}{2N}\sum_{i=1}^{N}\sum_{b\in\{A,B\}}
\ind[\hat y_i^b=y_i^b],\label{eq:acc}\displaybreak[1]\\
\pairacc&=\frac{1}{N}\sum_{i=1}^{N}\prod_{b\in\{A,B\}}
\ind[\hat y_i^b=y_i^b].\label{eq:pairacc}
\end{align}
Acc. credits individual answers; PairAcc requires both historical alternatives to be answered correctly. Unresolved answers count as incorrect.

\paragraph{Decomposing paired success.}
Let $n_{\mathrm{diff}}$ count pairs with two different valid predictions and $n_{\mathrm{both}}$ count pairs with both predictions correct. Define
\begin{equation}
D=\frac{n_{\mathrm{diff}}}{N},\qquad
C=\frac{n_{\mathrm{both}}}{n_{\mathrm{diff}}},\qquad
\pairacc=DC.
\label{eq:pair_decomposition}
\end{equation}
High $D$ indicates frequent differences in valid answers across histories; low $D$ indicates repeated or invalid answers. High $C$ means these differences usually answer both histories correctly; low $C$ means they often do not. High $C$ with low $D$ indicates correct discrimination on only a small subset. When $n_{\mathrm{diff}}=0$, $C$ is undefined and shown as ``--''.

\paragraph{Answer reversals.}
A pair is \emph{fully reversed} when $\hat y_i^A=y_i^B$ and $\hat y_i^B=y_i^A$. For $n_{\mathrm{rev}}$ such pairs, define
\begin{equation}
R=\frac{n_{\mathrm{rev}}}{N},\qquad S=\pairacc-R.
\label{eq:pair_reversal}
\end{equation}
High $R$ indicates frequent assignment of the correct answers to the opposite histories; low $R$ means such reversals are rare, not necessarily that answers are correct. Large positive $S$ indicates that correct pairings substantially outnumber reversals; negative $S$ indicates the opposite. Near zero $S$ can reflect either rare correct and reversed pairs or similar rates of both. Independent stochastic calls sharing an output distribution give $\mathbb{E}[S]=0$, regardless of option preferences. Thus answer changes alone need not reflect historical evidence use. We interpret $D$ and $S$ alongside identical input controls (see supplement).

\section{Benchmarking Existing Models}
\label{sec:models}
\subsection{Evaluation Settings}
Table~\ref{tab:our_benchmark} compares 34 configurations (A--E) and human references (F). Task scores measure branch accuracy; overall Acc. and PairAcc weight all 700 branches and 350 pairs equally. Detailed protocols are in the supplement.

\begin{table*}[t]
\centering
\caption{TwinPast results (\%). Task scores and Acc. measure branch accuracy; PairAcc requires both branches correct.}
\label{tab:our_benchmark}
\small
\setlength{\tabcolsep}{4pt}
\renewcommand{\arraystretch}{0.92}
\begin{tabular}{@{}l*{6}{r}>{\columncolor{black!8}}c*{4}{r}@{}}
\toprule
& \multicolumn{5}{c}{Historical tasks} & \multicolumn{2}{c}{Overall} & \multicolumn{4}{c}{Paired diagnostics} \\
\cmidrule(lr){2-6}\cmidrule(lr){7-8}\cmidrule(lr){9-12}
System / Model & AIR & HSR & EOR & HAC & MEI & Acc. & \textbf{PairAcc} & D & C & R & S \\
\midrule
\multicolumn{12}{l}{\textbf{(A) Online / Streaming Video LLMs}} \\
Dispider~\cite{qian2025dispider} & 64.67 & 54.55 & 63.85 & 43.86 & 34.75 & 53.86 & 19.43 & 25.43 & 76.40 & 2.00 & 17.43 \\
MOSS-VL Realtime~\cite{wang2026moss} & 82.07 & 56.49 & 66.15 & 40.35 & 38.14 & 59.29 & 32.86 & 41.43 & 79.31 & 1.43 & 31.43 \\
VideoLLM-online~\cite{chen2024videollm} & 45.65 & 48.05 & 46.92 & 34.21 & 23.73 & 40.86 & 0.86 & 1.71 & 50.00 & 0.29 & 0.57 \\
MiniCPM-o 4.5~\cite{cui2026minicpm} & 86.96 & 71.43 & 83.08 & 48.25 & 43.22 & 69.14 & 46.29 & 56.00 & 82.65 & 1.43 & 44.86 \\
Mage-VL (codec + gate)~\cite{yang2026mage} & 46.74 & 48.70 & 46.15 & 33.33 & 22.88 & 40.86 & 0.29 & 0.86 & 33.33 & 0.00 & 0.29 \\
JoyAI-VL Interaction~\cite{yao2026joyai} & 76.63 & 65.58 & 71.54 & 44.74 & 37.29 & 61.43 & 33.71 & 44.86 & 75.16 & 2.00 & 31.71 \\
StreamForest~\cite{zeng2025streamforest} & 83.15 & 62.99 & 67.69 & 47.37 & 42.37 & 63.14 & 35.43 & 44.29 & 80.00 & 1.43 & 34.00 \\
ReKV~\cite{di2025streaming} & 77.72 & 51.95 & 59.23 & 42.11 & 41.53 & 56.71 & 27.43 & 36.29 & 75.59 & 3.43 & 24.00 \\
OneStreamer~\cite{zeng2026onestreamer} & 66.30 & 61.69 & 67.69 & 38.60 & 38.14 & 56.29 & 26.29 & 38.00 & 69.17 & 2.29 & 24.00 \\
Qwen3.8-Omni-Flash-Realtime & 83.70 & 75.97 & 80.00 & 52.63 & 54.24 & 71.29 & 49.14 & 57.14 & 86.00 & 0.86 & 48.29 \\
AURA~\cite{lu2026aura} & 76.63 & 66.23 & 70.77 & 41.23 & 39.83 & 61.29 & 33.43 & 41.71 & 80.14 & 0.86 & 32.57 \\
\midrule
\multicolumn{12}{l}{\textbf{(B) Offline Video LLMs}} \\
Dispider & 65.22 & 54.55 & 64.62 & 43.86 & 35.59 & 54.29 & 21.43 & 26.86 & 79.79 & 1.14 & 20.29 \\
MOSS-VL Realtime & 85.87 & 68.83 & 76.15 & 44.74 & 44.92 & 66.71 & 44.86 & 53.14 & 84.41 & 1.43 & 43.43 \\
VideoLLM-online & 46.20 & 48.70 & 46.15 & 33.33 & 24.58 & 41.00 & 0.57 & 1.43 & 40.00 & 0.00 & 0.57 \\
MiniCPM-o 4.5 & 86.96 & 72.08 & 86.92 & 50.00 & 45.76 & 70.71 & 48.86 & 54.57 & 89.53 & 0.29 & 48.57 \\
Mage-VL (codec) & 84.24 & 68.83 & 81.54 & 52.63 & 49.15 & 69.29 & 46.29 & 58.00 & 79.80 & 1.71 & 44.57 \\
JoyAI-VL Interaction & 78.26 & 70.78 & 73.85 & 48.25 & 42.37 & 64.86 & 38.86 & 47.43 & 81.93 & 0.29 & 38.57 \\
AURA & 61.96 & 59.09 & 63.08 & 41.23 & 30.51 & 52.86 & 20.00 & 25.43 & 78.65 & 1.71 & 18.29 \\
DeepSeek-Flash (64 Frames) & 84.24 & 61.04 & 73.85 & 44.74 & 61.02 & 66.86 & 42.57 & 56.86 & 74.87 & 3.43 & 39.14 \\
Qwen3-VL-8B (64 Frames)~\cite{bai2025qwen3vltechnicalreport} & 82.61 & 60.39 & 60.00 & 39.47 & 44.07 & 60.00 & 34.00 & 48.29 & 70.41 & 4.57 & 29.43 \\
\midrule
\multicolumn{12}{l}{\textbf{(C) General purpose MLLMs}} \\
Gemini 4 Argon & 97.28 & 96.10 & 99.23 & 85.96 & 94.92 & 95.14 & 91.43 & 93.71 & 97.56 & 0.00 & 91.43 \\
Kimi K3.1 Pro & 98.37 & 94.16 & 99.23 & 82.46 & 91.53 & 93.86 & 89.43 & 93.43 & 95.72 & 0.57 & 88.86 \\
Seed 2.1 Pro & 95.65 & 94.81 & 93.08 & 85.09 & 90.68 & 92.43 & 86.57 & 91.43 & 94.69 & 0.29 & 86.29 \\
Seed 2.1 Turbo & 95.65 & 91.56 & 93.85 & 78.95 & 94.07 & 91.43 & 85.43 & 89.14 & 95.83 & 0.57 & 84.86 \\
Muse Spark 1.3 & 96.74 & 94.16 & 96.15 & 71.93 & 88.98 & 90.71 & 82.86 & 88.86 & 93.25 & 0.57 & 82.29 \\
Qwen3.8-Omni-Flash & 96.20 & 91.56 & 98.46 & 78.95 & 83.05 & 90.57 & 82.86 & 89.14 & 92.95 & 0.00 & 82.86 \\
MiniMax-M3.1-Flash & 95.11 & 90.91 & 93.85 & 79.82 & 79.66 & 88.86 & 80.57 & 86.57 & 93.07 & 0.29 & 80.29 \\
GLM-5.3-Flash & 93.48 & 83.77 & 94.62 & 57.02 & 76.27 & 82.71 & 71.43 & 83.43 & 85.62 & 2.57 & 68.86 \\
MiMo-V2.6-Flash & 90.22 & 83.12 & 93.08 & 49.12 & 68.64 & 78.86 & 66.57 & 78.57 & 84.73 & 1.71 & 64.86 \\
\midrule
\multicolumn{12}{l}{\textbf{(D) SimpleStream style recent window baselines}} \\
DeepSeek-Flash (1 Frame) & 46.20 & 48.70 & 41.79 & 36.26 & 25.71 & 40.86 & 5.14 & 15.90 & 32.34 & 4.95 & 0.19 \\
DeepSeek-Flash (4 Frames) & 44.57 & 49.35 & 40.77 & 34.21 & 31.36 & 41.00 & 3.14 & 13.14 & 23.91 & 3.71 & -0.57 \\
Qwen3-VL-8B (4 Frames) & 47.28 & 48.70 & 45.38 & 36.84 & 26.27 & 42.00 & 0.00 & 0.00 & -- & 0.00 & 0.00 \\
\midrule
\multicolumn{12}{l}{\textbf{(E) Blind LLMs (no visual input)}} \\
\blindicon\ DeepSeek-Flash (Query only) & 48.91 & 48.05 & 49.74 & 38.89 & 27.40 & 43.62 & 2.10 & 6.86 & 30.56 & 2.38 & -0.29 \\
\blindicon\ Qwen3-VL-8B (Query only) & 45.65 & 50.00 & 46.92 & 42.11 & 25.42 & 42.86 & 0.00 & 0.00 & -- & 0.00 & 0.00 \\
\midrule
\multicolumn{12}{l}{\textbf{(F) Human references}} \\
Human (offline replay) & 97.83 & 91.56 & 95.38 & 85.09 & 94.92 & 93.43 & 88.29 & 91.14 & 96.87 & 0.29 & 88.00 \\
Human (streaming) & 80.28 & 86.67 & 70.24 & 76.48 & 83.15 & 79.69 & 65.29 & 70.91 & 92.07 & 0.00 & 65.29 \\
\bottomrule
\end{tabular}
\end{table*}

\begin{figure*}[t]
\centering
\includegraphics[width=\textwidth,trim=0bp 6bp 0bp 8bp,clip]{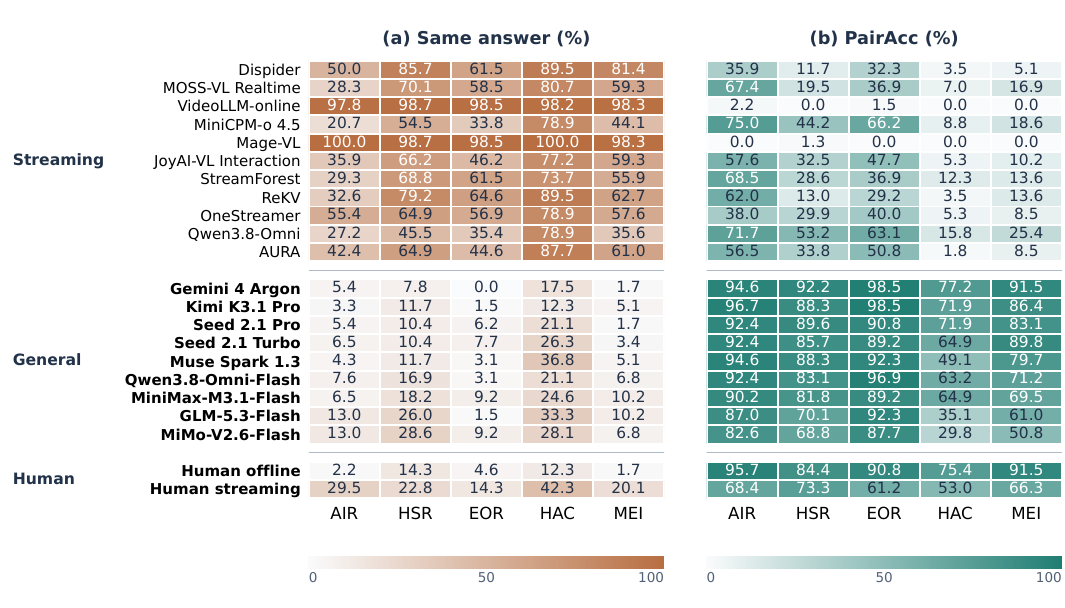}
\caption{Task wise paired answers: (a) same option; (b) both correct (PairAcc).}
\label{fig:paired_answer_patterns}
\end{figure*}

\noindent\textbf{(A) Online/streaming.} Adapted official inference code processes videos sequentially with each model's original memory; questions and options arrive only at query time. \textbf{(B) Offline.} Video and question are supplied together; most models receive up to 64 uniformly sampled frames.

\noindent\textbf{(C) General purpose.} Native video APIs receive each complete video with its question and options. \textbf{(D) Recent window.} Following SimpleStream~\cite{shen2026simplebaselinestreamingvideo}, models receive questions and options with the latest frame or four recent frames sampled at 1\,fps. \textbf{(E) Blind.} Models receive only questions and options.

\noindent\textbf{(F) Human references.} Offline replay allows two collaborators to inspect questions and replay videos. Streaming participants watch once without replay; questions and options appear at the annotated query time.

\subsection{What Paired Evaluation Reveals}
\paragraph{Historical input changes paired performance.}
Qwen3-VL-8B achieves 42.00\% Acc. with four recent frames but \textbf{answers no pairs correctly}. With 64 uniformly sampled frames, Acc. rises to 60.00\% and PairAcc to 34.00\%. DeepSeek-Flash shows the same pattern: Acc. increases from 41.00\% to 66.86\%, and PairAcc from 3.14\% to 42.57\%. Thus, recent inputs can support substantial individual accuracy while failing to distinguish the two histories. This reverses the input ranking in Table~\ref{tab:input_comparison}: recent frames outperform uniformly sampled history on OVO-Bench and StreamingBench, whereas historical input improves both Acc. and PairAcc on TwinPast. Holding recent context fixed makes the earlier differences necessary for consistently distinguishing the paired histories.

\paragraph{Streaming systems distinguish fewer histories correctly.}
All evaluated streaming systems score below the general purpose models in Table~\ref{tab:our_benchmark}(C). Their PairAcc ranges are 0.29--49.14\% and 66.57--91.43\%, respectively, leaving a 17.43 point gap between the strongest streaming system and the lowest scoring general purpose model. This gap persists within every historical task: the lowest PairAcc among the general purpose models exceeds the highest streaming PairAcc in all five groups (Fig.~\ref{fig:paired_answer_patterns}). The overall gap therefore extends across the evaluated historical capabilities.

\paragraph{Within family gaps in responses to history.}
Among models evaluated both online and offline, most achieve lower paired accuracy online. For example, MOSS-VL Realtime reaches 44.86\% PairAcc offline and 32.86\% online. The non-realtime and realtime variants of Qwen3.8-Omni achieve 82.86\% and 49.14\%, respectively. Their answer variation rates differ markedly: $D$ falls from 89.14\% to 57.14\%, while conditional correctness $C$ changes from 92.95\% to 86.00\%. The realtime variant answers both branches correctly on 86.00\% of pairs where it gives different valid answers, but produces such answers on far fewer pairs. Its lower paired performance is therefore accompanied by a much lower rate of different answers across histories: it correctly distinguishes a substantial subset of histories, but this success covers a smaller share of the dataset.

DeepSeek-Flash Query only and 4 Frames yield $D=6.86\%$ and $13.14\%$ on identical inputs, yet $S$ remains near zero. With 64 Frames, $S$ rises to 39.14 points ($R=3.43\%$).

\begin{samepage}
\section{Diagnosing Historical Evidence Use}
\label{sec:diagnostics}
We analyze failures in online historical QA by comparing paired answers and inspecting the evidence available at the final query.
\par
\end{samepage}

\subsection{Responses to Different Histories}
Counting (\textbf{HAC}) and multi-event integration (\textbf{MEI}) remain challenging: the highest PairAcc scores among the evaluated online systems are 15.8\% and 25.4\%, respectively (Fig.~\ref{fig:paired_answer_patterns}). Among the multi-event questions, 46 pairs use options that combine two binary facts. In 13 of these pairs, both facts change across the two branches. However, AURA's answers capture the change in only one fact, assigning the same value to the other in both branches. Thus, selecting a different option does not necessarily mean that the model has captured all changes in the historical facts. Further analyses for each task are provided in the supplementary material.

\subsection{Errors in Using Available Historical Evidence}
\begin{figure}[!tb]
\centering
\includegraphics[width=\columnwidth]{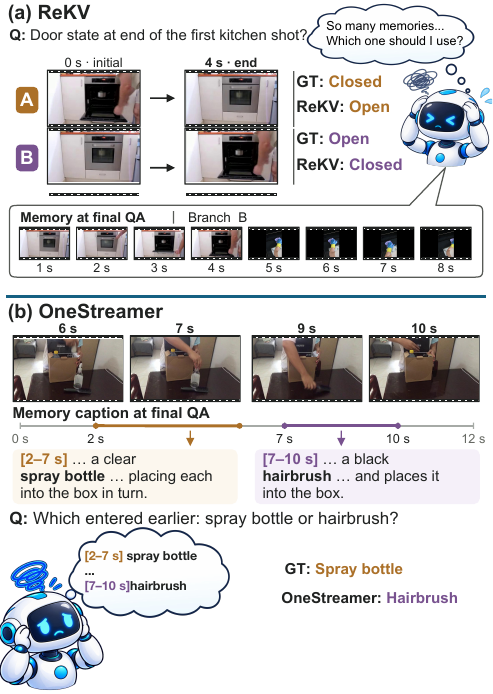}
\caption{Errors despite available evidence. (a) ReKV's answers match initial rather than queried states. (b) OneStreamer contradicts its recorded event order.}
\label{fig:evidence_use}
\end{figure}

In Fig.~\ref{fig:evidence_use}, ReKV is asked for the oven door state at the end of the first kitchen shot. Branch A shows an open-to-closed transition; branch B shows the reverse. All eight sampled frames per branch remain available at the final query, including the initial state at 0\,s and the queried state at 4\,s. ReKV answers \emph{open} for A and \emph{closed} for B. The answers differ across histories but match the initial states, forming a fully reversed pair. Correct historical discrimination requires associating each state with the time specified by the question.

OneStreamer exposes a gap between recording an event relation and using it to answer. Its text memory supplied to final QA records the spray bottle entering the box during 2--7\,s and the hairbrush during 7--10\,s. Asked which object entered earlier, it selects the hairbrush. The memory preserves both placements and their order, which supports the spray bottle as the answer. The final selection reverses the chronology supplied to the answering step.

The input comparisons establish the value of historical access. Here, both visual and textual memory contain the needed evidence, yet the answers conflict with it. Streaming historical QA therefore requires memory and retrieval to work with reasoning that connects the question to the relevant states and events, and derives the answer from their temporal context and relations.

\subsection{Beyond Narration: Following QA Instructions}
VideoLLM-online~\cite{chen2024videollm} illustrates a failure to use the instruction to determine which historical information to report. Asked how many complete claps occurred before a scene transition and instructed to select an option, it replies ``You clap your hands'' for both histories, whose counts are six and nine. The reply describes a shared activity while omitting the historical difference that determines the answer. Answering requires counting complete events before the transition and selecting the corresponding option. The response remains an activity description, leaving the requested count and selection unresolved.

This behavior highlights the need for task diversity in streaming VLM training. Describing an activity, counting its occurrences, and comparing event order require different uses of the same video. Alongside free-form narration, training should include QA instructions that vary the target facts, temporal scope, and required operation, with supervision for both answer content and format. Such supervision should teach models to select and interpret historical evidence according to the user's task. Further examples and free-generation results are in the supplement.

\section{Conclusion and Limitations}
TwinPast tests whether answers follow changes in historical evidence under shared recent context and text. Historical inputs improve performance on TwinPast, reversing the recent input advantage in the reported existing benchmark comparisons. Streaming systems lag behind general purpose video models across all five tasks. When the same model is evaluated in online and offline settings, paired accuracy is lower online in most cases; the realtime Qwen3.8-Omni variant likewise trails its non-realtime counterpart. Our cases show that retained evidence can still be used incorrectly: ReKV selects an initial rather than queried state, while OneStreamer contradicts event order recorded in its memory. Historical QA therefore requires both retaining relevant evidence and using it consistently with the requested time and temporal relations.

Our evaluation focuses on answering historical questions at specified query times. It does not assess proactive interaction, such as deciding when to initiate a response without an explicit query, or response latency; both are important in streaming video systems~\cite{wang2025videollm,wang2026mmduet2,ding2025streammind,Wang_2025_CVPR}. Future work can extend TwinPast's paired design to proactive interaction, where different histories determine whether and when to respond under shared recent observations. Joint evaluation of response content, timing, and latency would connect historical evidence use with timely proactive assistance.

\FloatBarrier
\clearpage
{\small\bibliographystyle{ieeenat_fullname}\bibliography{references}}
\end{document}